\documentclass{article}
\usepackage{arxiv}
\usepackage[utf8]{inputenc} 
\usepackage[T1]{fontenc}    
\usepackage[hyperfootnotes=false]{hyperref}       
\usepackage{url}            
\usepackage{booktabs}       
\usepackage{amsfonts}       
\usepackage{nicefrac}       
\usepackage{microtype}      
\usepackage{lipsum}		
\usepackage{graphicx}
\usepackage{grffile}
\usepackage[numbers]{natbib}
\usepackage{doi}
\usepackage{amsmath}
\usepackage{amsthm}
\usepackage{subcaption}
\usepackage{footnote}
\makesavenoteenv{table*}
\makesavenoteenv{tabular}
\usepackage{tablefootnote}
\usepackage{mwe}
\usepackage{float}
\usepackage{ragged2e}
\usepackage{multirow}
\usepackage{wrapfig}
\usepackage{xcolor}
\usepackage{setspace}
\floatstyle{plaintop}
\restylefloat{table}

\usepackage{etoolbox}

\title{CRAFT: Cultural Russian-Oriented Dataset Adaptation for Focused Text-to-Image Generation}

\author{\textbf{Viacheslav Vasilev}$^{1,2,*}$, \And \textbf{Vladimir Arkhipkin}$^1$, \And \textbf{Julia Agafonova}$^{1, 3}$, \And
\textbf{Tatiana Nikulina}$^1$, \And \textbf{Evelina Mironova}$^1$,
\And \textbf{Alisa Shichanina}$^1$, \And \textbf{Nikolai Gerasimenko}$^1$, \And
\textbf{Mikhail Shoytov}$^1$, \And
\textbf{Denis Dimitrov}$^{1, 4}$
}

\renewcommand{\shorttitle}{CRAFT: Cultural Russian-Oriented Dataset Adaptation for Focused Text-to-Image Generation}

\begin{document}
\maketitle

\begin{abstract}
Despite the fact that popular text-to-image generation models cope well with international and general cultural queries, they have a significant knowledge gap regarding individual cultures. This is due to the content of existing large training datasets collected on the Internet, which are predominantly based on Western European or American popular culture. Meanwhile, the lack of cultural adaptation of the model can lead to incorrect results, a decrease in the generation quality, and the spread of stereotypes and offensive content. In an effort to address this issue, we examine the concept of cultural code and recognize the critical importance of its understanding by modern image generation models, an issue that has not been sufficiently addressed in the research community to date. We propose the methodology for collecting and processing the data necessary to form a dataset based on the cultural code, in particular the Russian one. We explore how the collected data affects the quality of generations in the national domain and analyze the effectiveness of our approach using the Kandinsky 3.1 text-to-image model. Human evaluation results demonstrate an increase in the level of awareness of Russian culture in the model.
\end{abstract}

\keywords{Cultural Code, Russian Culture, Dataset Adaptation, Text-to-Image Generation, Fine-tuning, Captioning, Diffusion models, Data Engineering}

\section{Introduction}

The use of image generation models for various purposes through online applications and web editors has become the norm in modern daily life \cite{dalle3, Midjourney, podell2023sdxl, kastryulin2024yaartartrenderingtechnology, razzhigaev2023kandinsky, arkhipkin2024kandinsky30technicalreport, vladimir-etal-2024-kandinsky}. This increase in popularity and generation quality was made possible by training models on data from the Internet and large open datasets of text-image pairs, such as LAION-5B \cite{NEURIPS2022_a1859deb}, MS COCO \cite{mscoco}, etc. Much of this data allows models to learn a lot about international concepts, objects and events, and some elements of those national cultures that are most represented in modern mass media \cite{shankar2017classificationrepresentationassessinggeodiversity}. This raises the problem of a significant bias towards this data, often ignoring information that is represented in a smaller number of training examples. Such data domains can be the cultural characteristics of certain social groups, peoples and residents of different regions and states \cite{devries2019doesobjectrecognitionwork, gustafson2023pinpointingobjectrecognitionperformance}. As a result, such generative models do not take into account the cultural code of these groups well enough. Erroneous and incorrect generations in this case can lead to, at a minimum, an unsatisfactory result, limited applicability of the model and partial loss of interest among users, and, at a maximum, to unintentional insult, and the spread of stereotypes and social biases.

At the same time, it remains not entirely clear how to resolve the problem of cultural adaptation, since the concept of a cultural code is rather vague. To the best of our knowledge, there is currently no significant research in the field of visual content generation that would focus on working with the culture of a particular country or large enough community. The goal of this work is to analyze the problem of cultural awareness lack in text-to-image models and propose a solution that would move the field forward. We analyze the concept of a cultural code and describe a procedure for collecting a dataset that would allow us to adapt a generation model to better understand the concepts of Russian, Soviet and post-Soviet culture. We propose a \textbf{CRAFT} methodology (\textbf{C}ultural \textbf{R}ussian-Oriented Dataset \textbf{A}daptation for \textbf{F}ocused \textbf{T}ext-to-Image Generation), emphasizing the general versatility of our approach.

Therefore, the \textbf{contribution} of this article is as follows:
\begin{itemize}
    \item We expose the cultural awareness issues that current text-to-image generation models have. We discuss the concept of cultural code and identify domains of visual data that we believe should be taken into account for cultural adaptation of the model (Section \ref{sec:preliminaries}).
    \item We propose and describe a methodology for collecting and processing a cultural-aware dataset, in particular for Russian culture (Section \ref{sec:method}).
    \item We analyze the effectiveness of our approach and explore how the collected data affects the quality of Russian cultural data generation for the Kandinsky 3.1 model \cite{arkhipkin2024kandinsky30technicalreport, vladimir-etal-2024-kandinsky}. We report human evaluation results where we achieve higher quality compared to models lacking sufficient knowledge of Russian culture (Section \ref{sec:results}).
\end{itemize}

\begin{figure}
    \centering
    \includegraphics[bb=0 0 6563 4065, width=\linewidth]{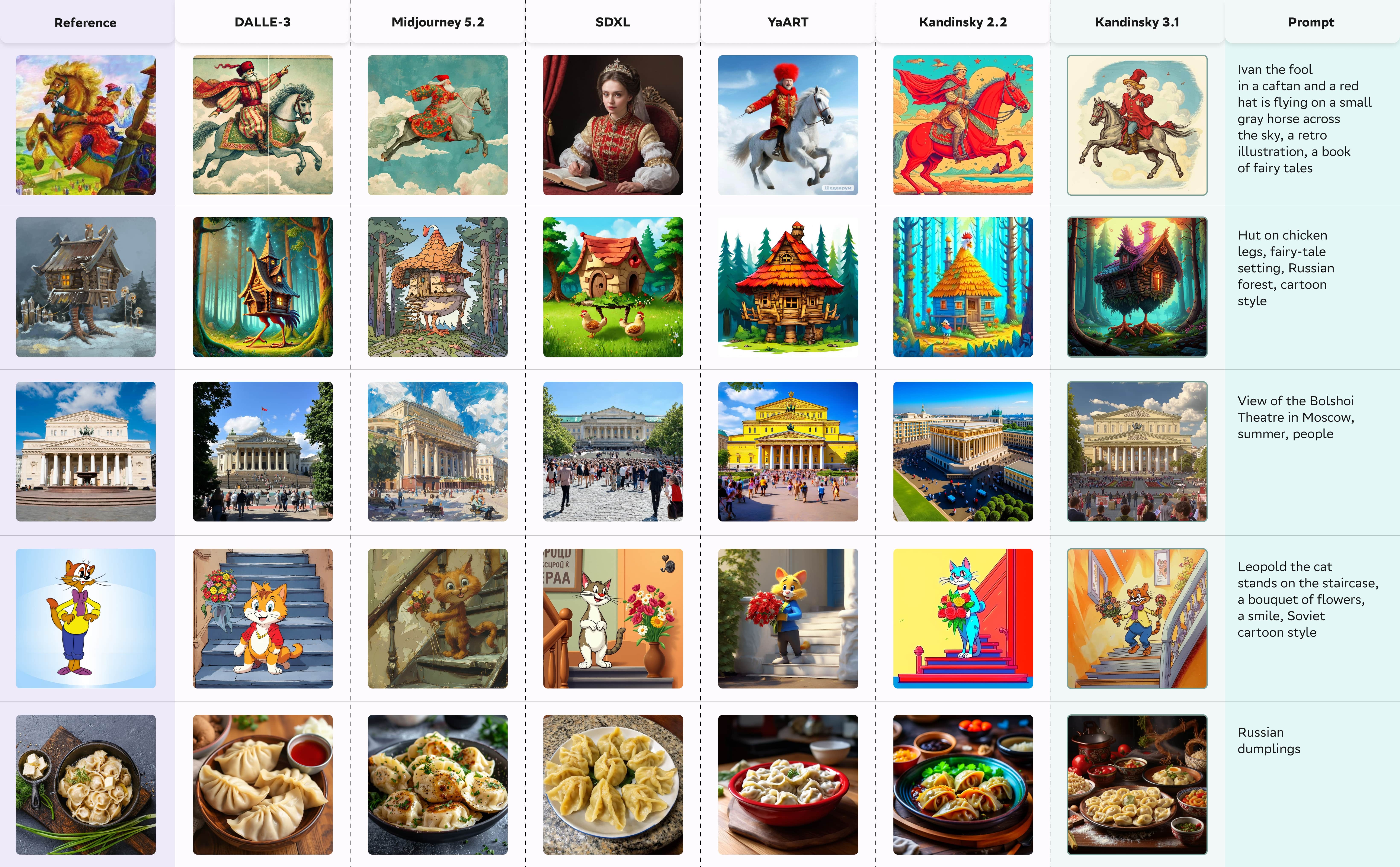}
    \caption{\textbf{Comparison of Russian cultural code generations for popular text-to-image models.} Reference is an example of a real image with a specific entity. The cultural adaptation procedure we propose helps improve the quality of cultural awareness for the Kandinsky 3.1 model, both in comparison with the previous version Kandinsky 2.2, and for other models.}
    \label{fig:comparison}
\end{figure}

\section{Related works}

{\centering
  \textit{2.1. Generative Models Finetuning}\par
}

Generative learning models in both natural language processing and computer vision are often pretrained on large datasets, allowing them to have extensive general knowledge about texts or images in the real world. Various finetuning methods, that is, additional training on a small dataset, can help expand the specific capabilities of the original models, both in terms of improving the generation quality and solving new types of problems \cite{Radford2018ImprovingLU, karras2020gans, nichol2022glide}. The key role here is not even the quantity of additional data, but its quality and detailed description. For example, the work \cite{dai2023emuenhancingimagegeneration} shows that finetuning, even on a small set of high-quality text-image data, can significantly improve the visual generation quality. Also, finetuning a small number of parameters with a relatively small number of images with detailed descriptions can significantly improve the quality of caption generation in visual-language models \cite{zhu2023minigpt4enhancingvisionlanguageunderstanding}. Additionally, finetuning a model on additional data is also a key element for model enhancement approaches such as domain adaptation \cite{10017290}, when the model is applied to data on which it was not pretrained, and AI alignment \cite{ji2024aialignmentcomprehensivesurvey}, when a certain behavior and understanding are expected from the model. In this work, we focus on preparing a high-quality dataset that would expand the text-to-image model's generation capabilities beyond the international visual domain and at the same time improve the visual quality of the generation of those national concepts that were already included in it at the pretraining stage. We also expect the model to expand its understanding of complex, culture-specific concepts, making it more relevant to the user who comes from a particular culture. In this sense, our work inherits the mentioned works.\newline

\begin{figure}
    \centering
    \includegraphics[bb=0 0 4468 2970, width=\linewidth]{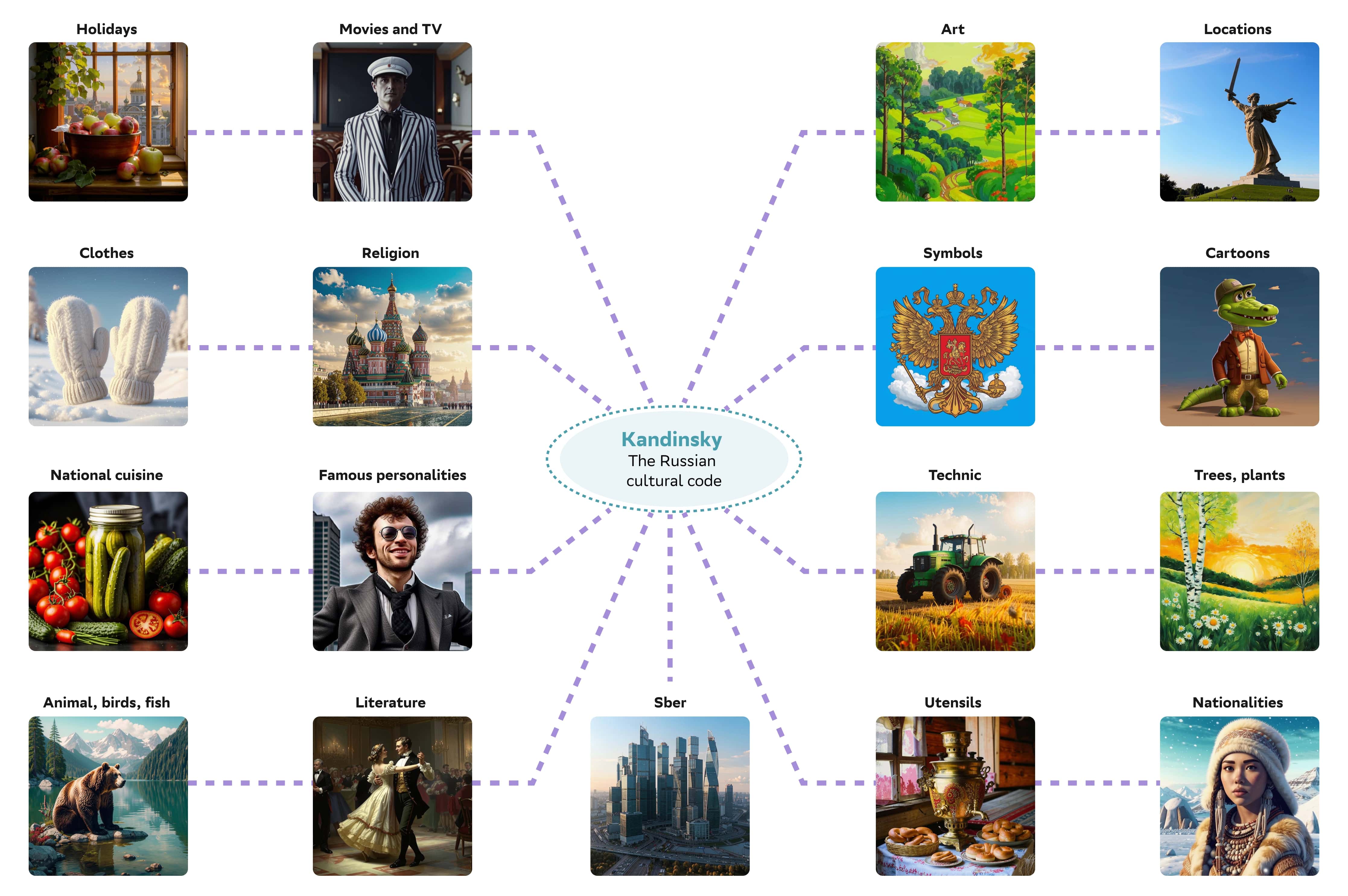}
    \caption{\textbf{17 main data categories for creating a dataset of the Russian cultural code.} Examples for each category are generated by the Kandinsky 3.1 model. For each category, we collect a set of visual entities from which we form our dataset. As a result of additional training, the model increases its level of cultural awareness.}
    \label{fig:categories}
\end{figure}

{\centering
  \textit{2.2. Cultural Adaptation}\par
}

By cultural awareness of a generation model we mean their ability to correctly resolve user requests and tasks that reflect certain features inherent in a particular culture. Cultural adaptation refers to techniques that allow models to achieve such results. Closely related to adaptation is the achievement of understanding linguistic and semantic features \cite{wibowo2023copal}, and the semantic matching of entities from different cultures \cite{10.1162/tacl_a_00634}, which has received attention in the adaptive translation \cite{peskov-etal-2021-adapting-entities}, offensive language detection \cite{zhou-etal-2023-cross, zhou-etal-2023-cultural, 10100717}, and cross-cultural natural language processing \cite{hershcovich-etal-2022-challenges}. Metaphor understanding and cultural awareness are also important for dialogue agents who interact directly with users \cite{cao-etal-2024-bridging}. The recent rise in the quality of multimodal architectures is driving interest in increasing cultural awareness for visual-language models (VLM) in question answering \cite{10.1145/3590773}, image-text retrieval and grounding \cite{bhatia2024localconceptsuniversalsevaluating}. Our work is one of the first to extend cultural awareness for text-based visual content generation.\newline

{\centering
  \textit{2.3. Ethics and Social Biases}\par
}

Lack of cultural awareness for visual generation models can lead to the spread of social bias and offensive content and pose risks comparable to those of using large language models (LLM) \cite{Katirai_2024}. Many works have been devoted to increasing the models' knowledge of distribution by race, skin color, gender, and geography, as well as analyzing and mitigating the social biases that were built into the model during pretraining stage \cite{9656762, 10.1145/3600211.3604711, NEURIPS2023_ae9500c4, 10.1145/3630106.3658968, Clemmer_2024_WACV}. Typically, such studies view cultural stereotypes as undesirable artifacts and negative phenomena, aiming for a higher level of globalization \cite{berg-etal-2022-prompt, 10.1613/jair.1.15388}. Although we agree with previous work on this point, we argue that avoiding culture-specific features may lead to even greater risks. In our work, on the contrary, we strive for cultural adaptation and enrichment of knowledge about cultural characteristics in order to reduce the excessive influence of globalization or Western-culture orientation, while maintaining correct behavior in terms of ethics and display of international concepts.

\section{Preliminaries}\label{sec:preliminaries}

{\centering
  \textit{3.1. Lack of Cultural Awareness of Text-to-Image Models}\par
}

The lack of cultural awareness and biases towards international or Western culture in generative models, resulting from the use of large datasets, has been noted previously \cite{bhatia2024localconceptsuniversalsevaluating, 10.1145/3600211.3604711, berg-etal-2022-prompt, 10.1613/jair.1.15388}. In this work, we focus on the performance of text-to-image models in the context of generating entities that have a close connection with Russian culture. To estimate the current level of Russian culture understanding by modern popular text-to-image models, we used relatively simple prompts that reflect the most popular concepts of mass culture, widely known in the post-Soviet states. We examine the performance of the DALL-E 3 \cite{dalle3}, Midjourney V5.2 \cite{Midjourney}, Stable Diffusion XL \cite{podell2023sdxl}, YandexART \cite{kastryulin2024yaartartrenderingtechnology} and Kandinsky 2.2 \cite{razzhigaev2023kandinsky} models. Figure \ref{fig:comparison} demonstrates that these models often fail to correctly generate the expected entities. It is worth noting that it is nevertheless impossible to talk about a complete lack of cultural awareness. Models more often generate examples of generalized concepts consisting of individual recognizable elements, but do not reflect the essence exactly as in the reference. We believe that even imprecise generation based on known elements is not a satisfactory representation of the desired entity, since such examples may also lead to critical reactions from users.\newline

\begin{figure}[t]
    \centering
    \includegraphics[bb=0 0 5868 1422, width=\linewidth]{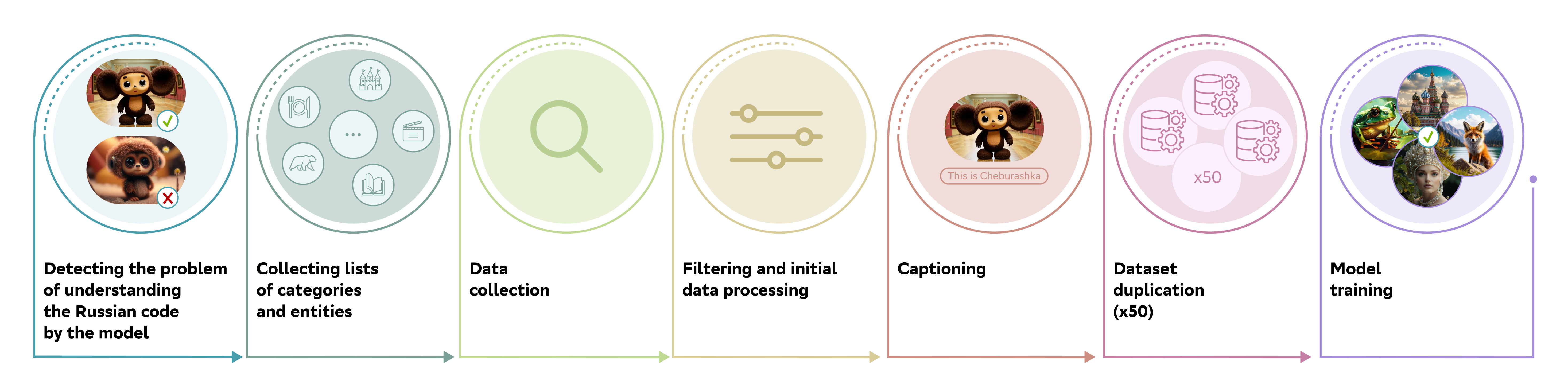}
    \caption{\textbf{General pipeline of our CRAFT method for cultural adaptation.} We create a list of categories and entities from Russian cultural code based on our own cultural analysis, collect and process data manually, including captioning process. The resulting dataset is used for additional training of the text-to-image model Kandinsky 3.1 \cite{arkhipkin2024kandinsky30technicalreport} to increase its level of cultural awareness.}
    \label{fig:pipeline}
\end{figure}

{\centering
  \textit{3.2. What is the Cultural Code?}\par
}

The cultural code is a complex concept at the intersection of cultural studies, sociology, philosophy, semiotics, communication theory and other sciences. A cultural code of a particular group of people is shaped by symbolic systems such as language, art, and tradition, as well as by the norms, values, social practices, and historical context. Media and visual types of information also play a significant role in the formation of the cultural code \cite{Corner1980}. We believe that in order for generative models to interpret or create culturally relevant content, they must be trained to understand these codes. This will significantly improve the visual quality of generation for certain prompts, develop the interpretive and communicative abilities of models, and allow the creation of systems that respect cultural differences, thereby avoiding biases and promoting more ethical views.

In this paper we consider a special case of the Russian cultural code. Following previous works in this field \cite{GOLOUBKOV2013107, billington2010icon, figes2002natasha, stites1992russian}, we consider language, literature, art, religion, philosophy, folk traditions and history as fundamental factors in the formation of Russian culture. Since the features of a particular culture are reflected in language \cite{https://doi.org/10.1525/eth.2002.30.4.401}, it is critical that the model understands metaphors, proverbs, and figurative expressions from everyday, colloquial speech. From the point of view of visual concepts, visual elements of modern popular culture, including cinema and television, and geographical and natural features are also important to us. In order to select data based on the results of our own cultural analysis, we created a list of 17 categories in which, we believe, the correct display of the visual cultural code is especially important. See Figure \ref{fig:categories} for the list of selected categories.

\begin{figure}[t]
    \centering
    \includegraphics[bb=0 0 3282 1745, width=0.75\linewidth]{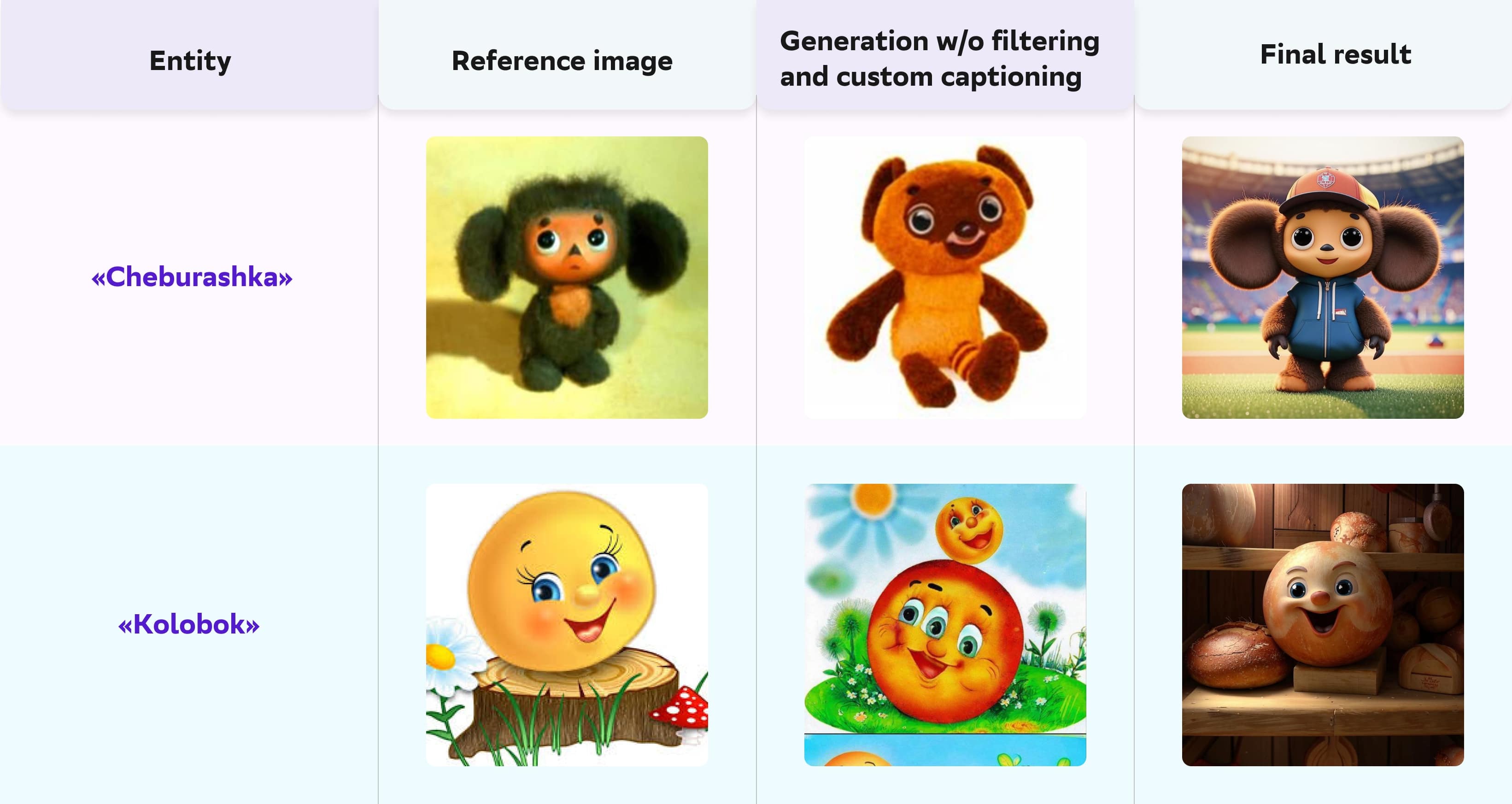}
    \caption{\textbf{The effect of filtering and custom captioning on the generation quality in Russian culture domain.} Reference image is a real image from the dataset that displays a specific entity. Our experiments showed that without additional data processing the model generates unsatisfactory results.}
    \label{fig:ablation}
\end{figure}

\section{Cultural Dataset Adaptation}\label{sec:method}

In this section, we describe the procedure for collecting and processing the dataset necessary for the cultural adaptation of the text-to-image model Kandinsky 3.1 \cite{arkhipkin2024kandinsky30technicalreport} in the context of the Russian cultural code. Figure \ref{fig:pipeline} shows the general scheme of the cultural adaptation process, each component of which we discuss in detail below.\newline

{\centering
  \textit{4.1. Entities Determination \& Data Collection}\par
}

In this work, by ``entity'' we mean a specific concept or object belonging to the category of cultural code. For example, the entities in the category \texttt{``trees''} are \texttt{``spruce''}, \texttt{``birch''}, \texttt{``pine''}, \texttt{``oak''}, etc. Categories such as \texttt{``literature''} and \texttt{``art''} include the titles of books, paintings, literary characters, as well as the names of famous writers, poets, artists, composers, and the like. For each category of Russian culture from the list we have defined, we have compiled a table of the most popular entities.

Since we collected data from open sources on the Internet, for each entity we also defined a set of web queries in Russian, which we automatically searched for relevant and correct data related to each entity. Thus, we manually selected about 8 thousand entities. Examples of entities from our Russian cultural code dataset and a set of web queries to them, translated into English, can be seen in Table \ref{tab:entities}.\newline

\begin{table}[!t]
\caption{Examples of entities from different categories with corresponding web queries.}   
\centering
\small
\begin{tabular}{ll}
    \hline
    \textbf{Entity} & \textbf{Web queries} \\
    \hline
    \textbf{Famous personalities} & \\
    \hline
    Yuri Gagarin & \texttt{Yuri Alekseyevich Gagarin, cosmonaut Gagarin,}\\
    & \texttt{Gagarin in a helmet with the inscription USSR, Gagarin in an orange spacesuit}\\
    \hline
    \textbf{Movies and TV} & \\
    \hline
    Baron Munchausen & \texttt{Baron Munchausen from ``The Very Same Munchhausen''},\\
    & \texttt{Baron Karl Friedrich Hieronymus von Münchhausen},\\ 
    & \texttt{actor Oleg Ivanovich Yankovsky, film ``The Very Same Munchhausen''} \\
    \hline
    \textbf{Literature} & \\
    \hline
    Maxim Gorky & \texttt{Alexei Maximovich Peshkov, Maxim Gorky}, \\
    & \texttt{Alexei Maximovich Gorky, Jehudiel Khlamida} \\
    \hline
    \textbf{Holidays} & \\
    \hline
    Christmas & \texttt{Nativity of Christ January 7 Orthodox holiday}\\
    \hline
    \textbf{National cuisine} & \\
    \hline
    Pirozhok & \texttt{pirozhok with cabbage, pirozhok with meat, pirozhok with apples}\\
    \hline
    \textbf{Utensils} & \\
    \hline
    Samovar & \texttt{Samovar, Russian samovar, ancient samovar, samovar-egg, samovar-flame},\\
    & \texttt{samovar-jug, samovar-ball, samovar-rooster, samovar-barrel, samovar-jar,}\\
    & \texttt{samovar-vase, samovar-dula, samovar-pear, samovar-glass,}\\
    & \texttt{acorn samovar, store samovar, travel samovar}\\
    \hline
    \textbf{Technic} & \\
    \hline
    VAZ-2101 & \texttt{VAZ-2101, ``Zhiguli'', ``Zhiga'', ``Zhigul'', ``Kopeyka'', ``Yedinichka'', ``Odnerka''} \\
    \hline
    \textbf{Trees and plants} & \\
    \hline
    Birch & \texttt{Birch, Betula pubescens, downy birch, Betula platyphylla},\\
    & \texttt{Asian white birch, Betula raddeana, weeping birch, warty birch, Betula utilis}\\
    \hline
\end{tabular}
\label{tab:entities}
\end{table}

{\centering
  \textit{4.2. Data Processing}\par
}

We found that training the model on data collected from the Internet according to our entity lists produces unsatisfactory results (Figure \ref{fig:ablation}). This is explained by the fact that a large number of entities correspond to a small number of examples, which, moreover, have low visual quality. Such examples include characters from old cartoons, movies, as well as old photographs, illustrations, some paintings and portraits. In order to solve this problem, we manually filtered the collected data, paying attention to the visual aesthetic of the images and their belonging to the Russian cultural code. The key points in the instructions for filtering by culture code are as follows:

\begin{itemize}
    \item If the images include elements of the cultures of other nations and nationalities from post-Soviet countries, then such data is acceptable;
    \item If the images contain something modern or partly not belonging to Russian culture, but is present along with elements of the Russian cultural code, then such data is acceptable;
    \item If the frame is from a film or TV series of non-Russian production, but it contains elements of Soviet and Russian culture, then such data is acceptable;
    \item If the general appearance of the scene does not give clues as to whether there are actually elements of Russian cultural code there, then such a scene is not acceptable.
\end{itemize}

Key points for quality filtering:

\begin{itemize}
    \item Poor quality images (blurry, overexposed, low sharpness, or homemade) are not accepted. At the same time, examples when good quality data does not exist for a particular entity should be processed separately;
    \item Images must not contain watermarks;
    \item Entities must not be distorted.
\end{itemize}

\begin{figure}[!ht]
    \centering
    \includegraphics[bb=0 0 2256 942, width=0.8\linewidth]{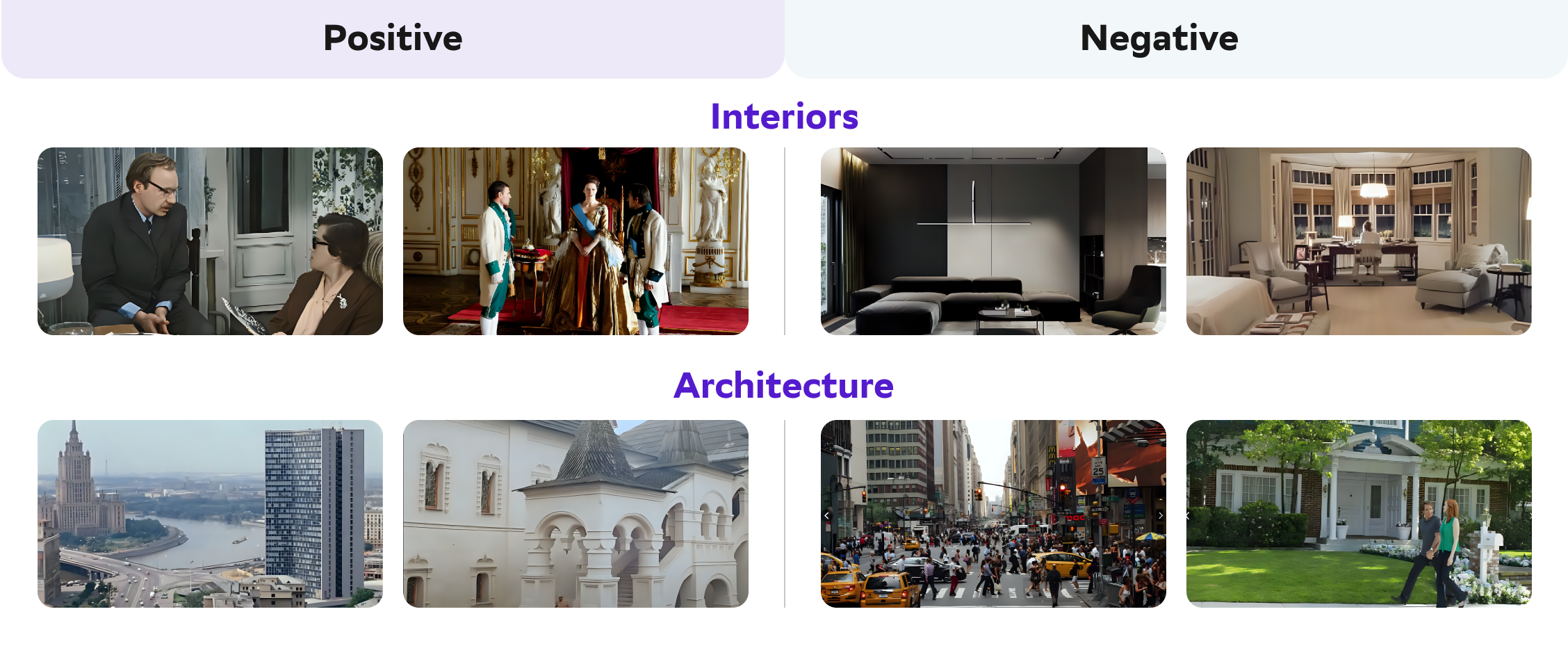}
    \caption{\textbf{Examples of images from the instructions that people followed when data filtering.} The task was to select only those images that contain Russian and post-Soviet visual features.}
    \label{fig:markup}
\end{figure}

Each estimator got acquainted with negative and positive examples of which data should be selected. Examples of such images are shown in Figure \ref{fig:markup}.\newline

\begin{table}
     \begin{center}
     \caption{Comparison of captions created by LLaVa-Next 34B \cite{liu2023improved} or people. The model quite often makes factual mistakes and does not know the titles and proper names.}
     \begin{tabular}{ p{3cm}  p{7cm}  p{5cm}  }
     \toprule
      Image & LLaVa-Next 34B & Human \\ 
     \hline
     \raisebox{-\totalheight}{\includegraphics[bb=0 0 986 558, width=0.3\textwidth]{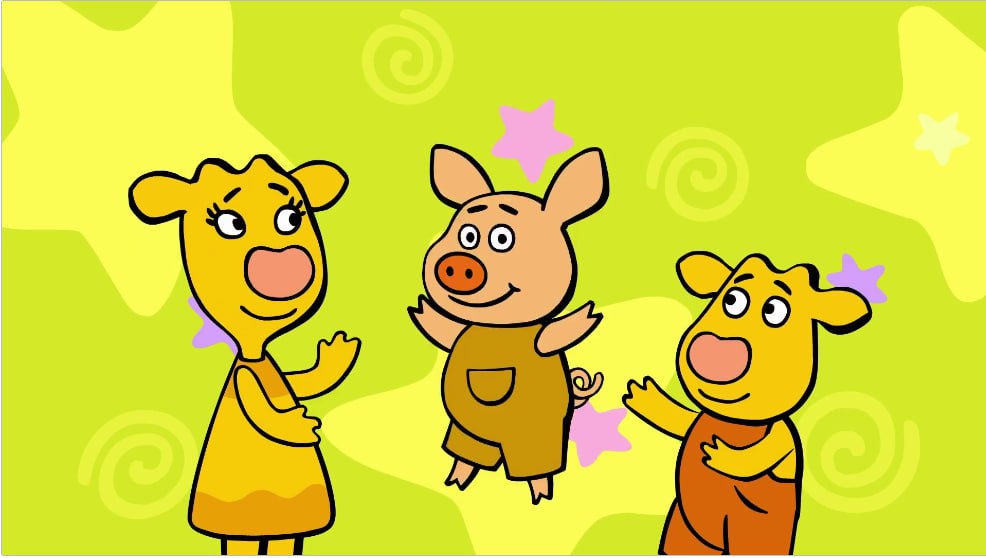}}
      & 
      \texttt{\scriptsize{\begin{spacing}{0.5}Cartoon {\color{red}character} that looks like a pig. The pig is drawn in a simplistic, child-friendly style, with bright outlines and flat colors. {\color{red}She} has a large round head with two small black eyes and a small {\color{red}triangular nose}. The pig's mouth is open in a smile, {\color{red}showing one tooth}, he has two small ears and {\color{red}a tuft of hair on the top of his head}, the pig is wearing a brown {\color{red}collared shirt}, the background is {\color{red}a solid color}, bright green, the overall style is cheerful and cartoonish, likely aimed at a younger audience.\end{spacing}}}
      & 
      \texttt{\scriptsize{\begin{spacing}{0.5}A still from the cartoon {\color{green}``Orange Cow'': Bo} the calf in an orange dress with a dark border and {\color{green}Zo} the calf in an orange-brown overalls look in surprise at the piglet. The pig is drawn in a simplified style: it has a large round head with two small black eyes, a small round nose and {\color{green}a cute smile. The pig is wearing a brown suit.} The background is {\color{green}bright green with light-colored curls, large yellow stars and small pink ones.}\end{spacing}}}
      \\ 
      \hline
      \raisebox{-\totalheight}{\includegraphics[bb=0 0 989 742, width=0.3\textwidth]{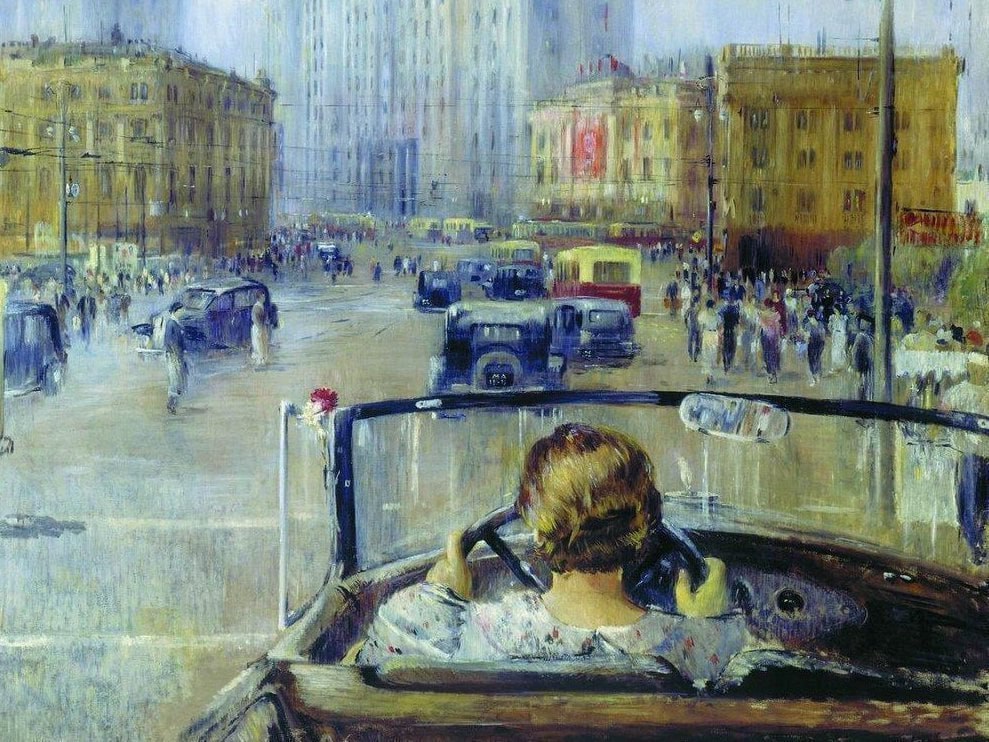}}
      & 
      \texttt{\scriptsize{\begin{spacing}{0.5}The scene is teeming with life, with numerous figures and vehicles populating the street. In the foreground, a woman is seated in a car, her back to the viewer, adding a sense of depth and perspective to the composition. The car is a prominent feature, its details rendered with a certain level of precision, contrasting with the more abstract background.The background is a flurry of activity, with buildings, cars, and people filling the canvas. The buildings, painted in warm hues, tower over the scene, their details softened by the artist's brushstrokes. The cars, rendered in cooler tones, add a dynamic element to the composition, their movement suggested through the artist's use of color and line. {\color{red}The color palette is dominated by warm tones, with the buildings and vehicles bathed in a soft, diffused light.} The artist has used a variety of brushstrokes, from loose and fluid to more defined and structured, creating a sense of movement and energy.Overall, the painting is a beautiful representation of an urban landscape, capturing the essence of city life through the lens of impressionism. The artist's use of color, light, and brushwork brings the scene to life, creating a sense of movement and energy that is both captivating and engaging.\end{spacing}}}
      & 
      \texttt{\scriptsize{\begin{spacing}{0.5}{\color{green}Painting by Soviet artist Yuri Pimenov ``New Moscow''}, which depicts a girl with short hair, wearing blue clothes with a white collar with long sleeves, sitting behind the wheel, which she holds with both hands, in an open-top car, back to her shoulders; a narrow windshield in a dark frame; a distant view mirror is attached to the right side of the glass. {\color{green}To the left of the car is a gray road}; in the upper left corner you can see a four-story yellow building, as well as blue cars and people crossing the road. In the upper right corner there are two four-story yellow buildings, next to which there are blue cars and red-yellow buses, and people are walking along the sidewalk on the right.\end{spacing}}}
      \\
      \hline
      \raisebox{-\totalheight}{\includegraphics[bb=0 -100 960 1280, width=0.3\textwidth]{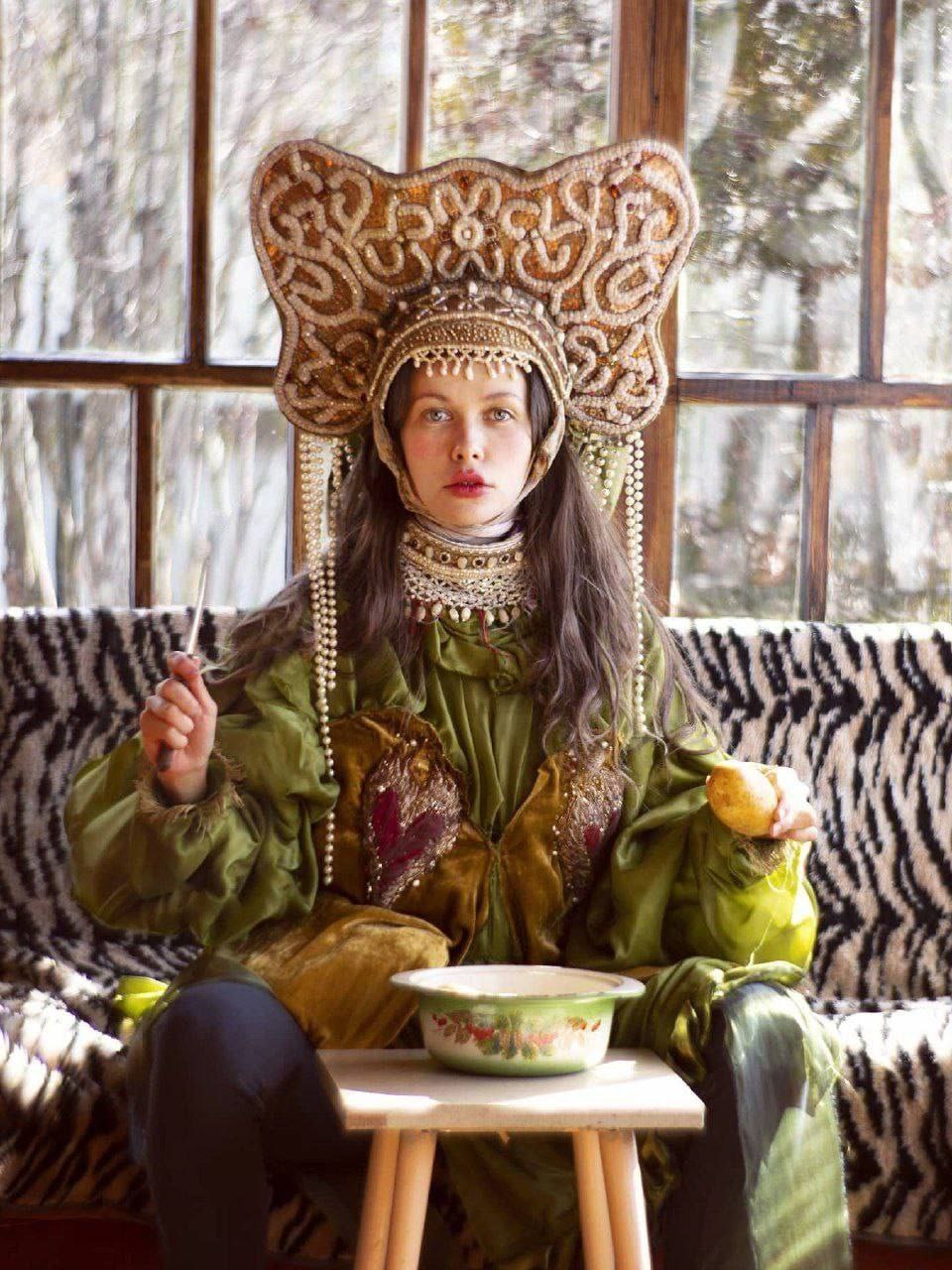}}
      & 
      \texttt{\scriptsize{\begin{spacing}{0.5}The image depicts a {\color{red}person} seated in a room with a large window in the background, through which a blurred view of trees can be seen. The individual is wearing a costume that appears to be inspired by historical or fantasy attire. {\color{red}The costume includes a headpiece with intricate patterns and what seems to be a beaded or embroidered veil that covers the head and neck.} The person's face is visible, and they are looking directly at the camera with a neutral expression.The costume consists of a green, long-sleeved garment with a high collar and a patterned chest area that includes what looks like a heart or a {\color{red}floral motif}. The person is holding a small, round object in their right hand, which could be a fruit or a decorative item.The room has a rustic feel, with a wooden floor and a zebra print sofa or bench on which the person is seated. The table in front of the person holds a bowl with a floral pattern, suggesting a setting for a meal or a tea. {\color{red}The overall style of the image is a blend of historical fashion photography with a touch of fantasy or costume drama.} The lighting is soft and natural, enhancing the textures of the costume and the room's ambiance.\end{spacing}}}
      & 
      \texttt{\scriptsize{\begin{spacing}{0.5}Photo of {\color{green}Uldus Bakhtiozina} in an outfit of his own design. The girl is wearing a light {\color{green}green silk dress}, on the front of which there is a yellow insert with hearts on the chest. On the girl’s head is a large golden {\color{green}kokoshnik} with light patterns, decorated with {\color{green}pearls}. The girl sits on a black and white striped sofa, behind which there is a large wooden window, holding a knife in her right hand, a potato in her left, and in front of her there is a green saucepan on a stool.\end{spacing}}}
      \\
      \bottomrule
      \end{tabular}
      \label{table:captioning}
      \end{center}
\end{table}

{\centering
  \textit{4.3. Captioning}\par
}

\textbf{Automatic captioning.} After the initial filtering, we ran automatic captioning for the collected data using the  LLaVa-NeXT model \cite{liu2023improved}. To obtain the high quality captions, we varied the following hyperparameters:

\begin{itemize}
    \item \textit{Maximum number of output tokens:} from 100 to 1024. Setting a small value resulted in a truncated description, or significant elements of the image were not revealed. We chose 512 tokens empirically.
    \item \textit{Temperature.} When the temperature was set to high (0.7-1), the model hallucinated and described objects that were not in the image, or concentrated on insignificant elements.
    \item \textit{Instructional prompt.} We used 3 options: short (\texttt{``Describe this image shortly in 1-2 short sentences''}), medium (\texttt{``Please provide a caption for this image. Speak confidently and describe everything clearly. Do not lie and describe only what you can see''}) and long (\texttt{``Describe this image and its style in a very detailed manner''}). A short prompt did not allow generating meaningful captions, and a long one described non-existent objects. Thus, we chose the middle option.
\end{itemize}

The generated captions were cleaned using regular expressions from service phrases and repeated initial sentences. However, we have found that such captions often contain factual errors and do a poor job of understanding the Russian cultural code. See Table \ref{table:captioning} for examples. We conducted time measurements that showed that people spend an average of $4.52$ minutes to create their own captions, and $5.23$ minutes on editing a synthetic caption. Therefore, we have decided to use only human-created captions.

\textbf{Manual captioning.} Empirically, we decided to create detailed captions from 2 to 10 sentences in length and according to the following instructions:

\begin{itemize}
    \item The greatest attention should be paid to the main entity;
    \item It is necessary to add the name of the character, the title of the book, movie, etc;
    \item Captions should be varied and not stick to one template;
    \item Phrases that suggest viewing an image should be avoided. For example, \texttt{``we see in this photo''}, \texttt{``this image shows''}, etc;
    \item It is allowed to use phrases that characterize the features of the image. For example, \texttt{``black and white photo''}, \texttt{``still from a Soviet cartoon''}, etc;
    \item Do not create descriptions that contain guesses about what this object in the image can be used for;
    \item Do not use imprecise language. For example, \texttt{``the people in the image are probably young''};
    \item It is necessary to skip collage images and images that contain text.
\end{itemize}

All created captions were reviewed by another group of people according to the same instructions. In this way, we collected about 200 thousand text-image pairs. To train the model, all captions were translated into English.\newline

{\centering
  \textit{4.4. Data Duplication \& Model Finetuning}\par
}

At the final stage, we duplicated the dataset 50 times to expand those domains and entities for which there are only a small number of high-quality training examples. Kandinsky 3.1 \cite{arkhipkin2024kandinsky30technicalreport} finetuning took place in two stages. The first is in $768 \times 768$ resolution with batchsize $=4$, and the second is in a mixed resolution of $768^2 \leq W \times H \leq 1024^2$ with a batchsize $=1$. The total number of steps is 500 thousand on 416 A100 GPU.

\section{Evaluation}\label{sec:results}

\begin{figure}
  \centering
  \includegraphics[bb=0 0 1088 224, width=\linewidth]{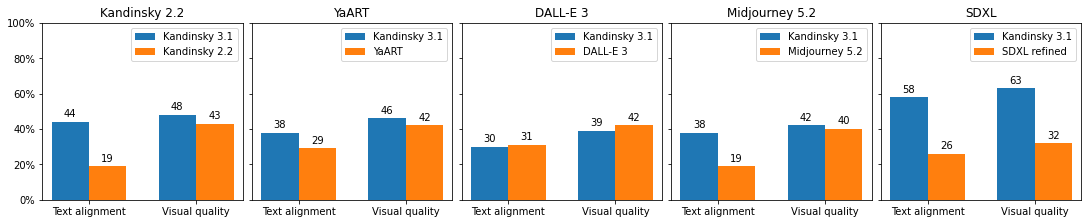}
  \caption{\textbf{Human evaluation results.}}
  \label{fig:sbs_results}
\end{figure}

Since there are no benchmarks and automatic metrics for the task of Russian cultural adaptation of text-to-image generation, we focused on conducting side-by-side human evaluation and collected our own set of 120 prompts, reflecting the features of the Russian cultural code. 12 people took part in the comparisons and each chose the best image according to the correspondence of the image content to the text prompt (text alignment) and the visual quality of the image.

The results are presented in the Figure \ref{fig:sbs_results}. As can be seen, Kandinsky 3.1 significantly outperforms the previous version Kandinsky 2.2 \cite{razzhigaev2023kandinsky}, Midjourney 5.2 \cite{Midjourney}, SDXL \cite{podell2023sdxl} and Russian model YaART \cite{kastryulin2024yaartartrenderingtechnology}. In comparison with the DALL-E 3 model \cite{dalle3}, we demonstrate competitive results, the difference in which is small and can fluctuate depending on test prompts and categories. The results of model generation before and after finetuning in Russian culture domain are presented in \hyperlink{appendix}{Appendix}.

\section{Limitations}

\begin{figure}[t]
\centering
\begin{minipage}[h]{\linewidth}
\center{\includegraphics[bb=0 0 12753 3976, width=1\linewidth]{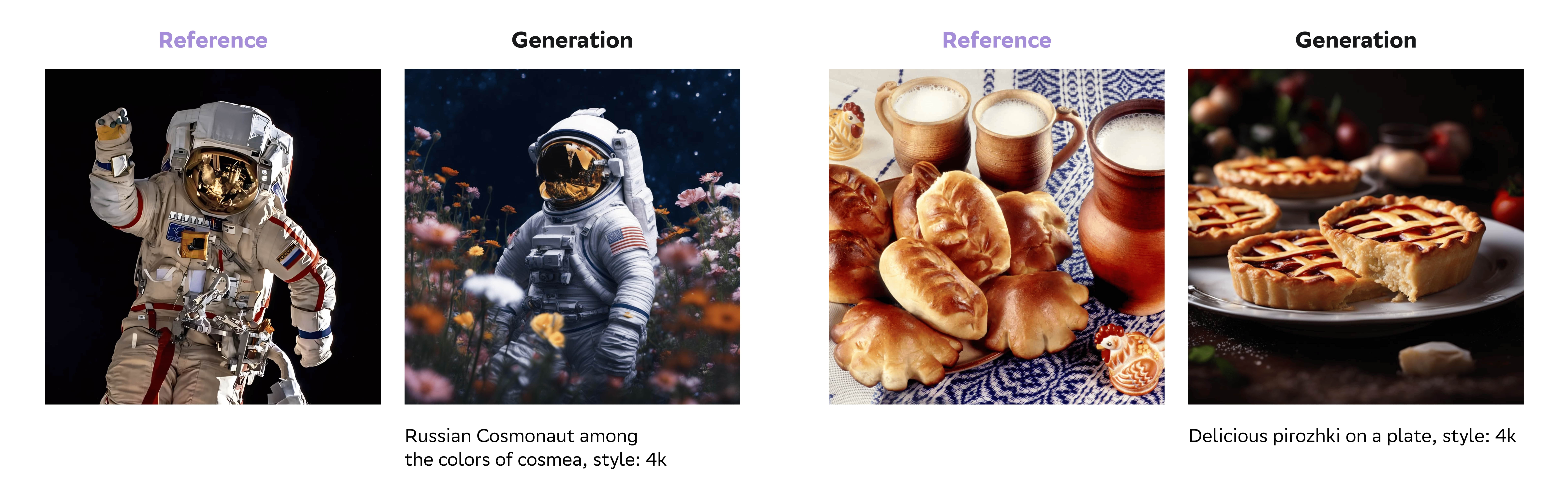}} \\
\end{minipage}
\vfill
\begin{minipage}[h]{\linewidth}
\center{\includegraphics[bb=0 0 12753 3976, width=1\linewidth]{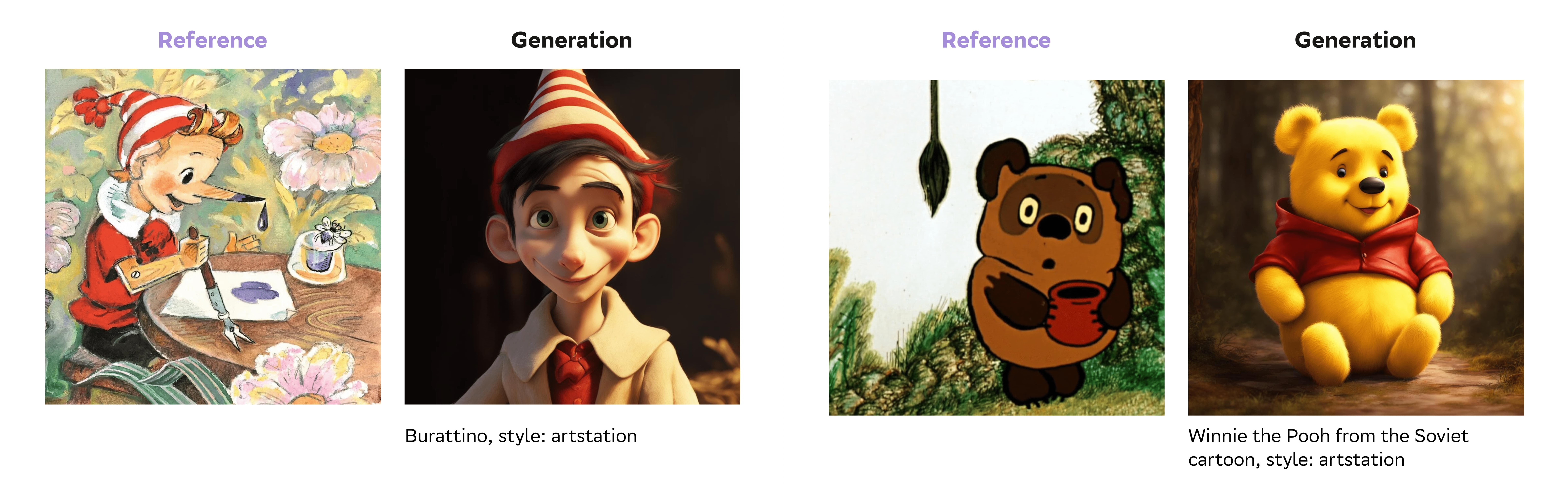}} \\
\end{minipage}
\caption{\textbf{Examples of incorrect generation for some entities.} Problems arise due to entities imbalance and translation difficulties.}
\label{fig:limitations}
\end{figure}

Currently, the generation in the domain of cultural code has problems for those entities that are represented by a small number of samples. For example, almost all available data on the appearance of characters in old films or cartoons are low-quality frames. Almost all data on the Internet related to these entities are copies of these frames.

The second problem is translation difficulties. For example, due to the fact that the Russian word ``cosmonaut'' is often translated into English as ``astronaut'', most of the photographs of cosmonauts in the dataset are represented with the American flag. The same applies to examples like ``pirozhki-pies'', and other (Figure \ref{fig:limitations}).

\section{Conclusion}

In this work, we are one of the first to propose an approach to enhance the cultural awareness of the text-to-image generation model. We collected a dataset of Russian cultural code and used it to improve the generation quality in this domain for our Kandinsky 3.1 model. Human evaluation of generation quality demonstrated the advantage of the described methodology.

We believe that our approach deserves further development. First of all, we strive to conduct an even deeper cultural analysis in order to expand the number of entities and data, including the semantic complexity of the test dataset. Our next priorities include applying advanced augmentation and entity balancing techniques for the dataset, using RAG \cite{NEURIPS2020_6b493230, zhao2024retrievalaugmentedgenerationaigeneratedcontent} for cultural adaptation, and extending our approach to text-to-video generation \cite{arkhipkin2023fusionframesefficientarchitecturalaspects, 10815947}. We also pay close attention to the ethical side of cross-cultural awareness and plan to develop advanced approaches to limit incorrect queries while still leaving a sufficient level of freedom for users.

\section{Later update}

This research continues in our next paper, where we propose a benchmark for evaluating the level of Russian cultural code understanding for text-to-image generation models \cite{vasilev-etal-2025-ruscode}. For more details, please refer to the publication.

\section*{Acknowledgments}

The authors express special gratitude to Igor Pavlov and Anastasia Lysenko, without whom this work would not have been done. We also thank Sergey Markov, his team from SberDevices, and the prompt-engineering and markup team: Anastasia Alyaskina, Nadezhda Martynova, Marina Yakushenko, Margarita Geberlein, Denis Kondratev, Stefaniya Kozlova, Mikhail Kornilin, and Alexandra Averina.

\bibliographystyle{unsrt}
\bibliography{references}

\begin{thebibliography}{10}

\bibitem{dalle3}
James Betker, Gabriel Goh, Li~Jing, Tim Brooks, Jianfeng Wang, Linjie Li, Long Ouyang, Juntang Zhuang, Joyce Lee, Yufei Guo, Wesam Manassra, Prafulla Dhariwa, Casey Chu, Yunxin Jiao, and Aditya Ramesh.
\newblock Improving image generation with better captions, 2023.

\bibitem{Midjourney}
Midjourney.
\newblock Midjourney.
\newblock \url{https://www.midjourney.com/}.

\bibitem{podell2023sdxl}
Dustin Podell, Zion English, Kyle Lacey, Andreas Blattmann, Tim Dockhorn, Jonas Müller, Joe Penna, and Robin Rombach.
\newblock Sdxl: Improving latent diffusion models for high-resolution image synthesis, 2023.

\bibitem{kastryulin2024yaartartrenderingtechnology}
Sergey Kastryulin, Artem Konev, Alexander Shishenya, Eugene Lyapustin, Artem Khurshudov, Alexander Tselousov, Nikita Vinokurov, Denis Kuznedelev, Alexander Markovich, Grigoriy Livshits, Alexey Kirillov, Anastasiia Tabisheva, Liubov Chubarova, Marina Kaminskaia, Alexander Ustyuzhanin, Artemii Shvetsov, Daniil Shlenskii, Valerii Startsev, Dmitrii Kornilov, Mikhail Romanov, Artem Babenko, Sergei Ovcharenko, and Valentin Khrulkov.
\newblock Yaart: Yet another art rendering technology, 2024.

\bibitem{razzhigaev2023kandinsky}
Anton Razzhigaev, Arseniy Shakhmatov, Anastasia Maltseva, Vladimir Arkhipkin, Igor Pavlov, Ilya Ryabov, Angelina Kuts, Alexander Panchenko, Andrey Kuznetsov, and Denis Dimitrov.
\newblock Kandinsky: an improved text-to-image synthesis with image prior and latent diffusion, 2023.

\bibitem{arkhipkin2024kandinsky30technicalreport}
Vladimir Arkhipkin, Andrei Filatov, Viacheslav Vasilev, Anastasia Maltseva, Said Azizov, Igor Pavlov, Julia Agafonova, Andrey Kuznetsov, and Denis Dimitrov.
\newblock Kandinsky 3.0 technical report, 2024.

\bibitem{vladimir-etal-2024-kandinsky}
Arkhipkin Vladimir, Viacheslav Vasilev, Andrei Filatov, Igor Pavlov, Julia Agafonova, Nikolai Gerasimenko, Anna Averchenkova, Evelina Mironova, Bukashkin Anton, Konstantin Kulikov, Andrey Kuznetsov, and Denis Dimitrov.
\newblock Kandinsky 3: Text-to-image synthesis for multifunctional generative framework.
\newblock In Delia~Irazu Hernandez~Farias, Tom Hope, and Manling Li, editors, {\em Proceedings of the 2024 Conference on Empirical Methods in Natural Language Processing: System Demonstrations}, pages 475--485, Miami, Florida, USA, November 2024. Association for Computational Linguistics.

\bibitem{NEURIPS2022_a1859deb}
Christoph Schuhmann, Romain Beaumont, Richard Vencu, Cade Gordon, Ross Wightman, Mehdi Cherti, Theo Coombes, Aarush Katta, Clayton Mullis, Mitchell Wortsman, Patrick Schramowski, Srivatsa Kundurthy, Katherine Crowson, Ludwig Schmidt, Robert Kaczmarczyk, and Jenia Jitsev.
\newblock Laion-5b: An open large-scale dataset for training next generation image-text models.
\newblock In S.~Koyejo, S.~Mohamed, A.~Agarwal, D.~Belgrave, K.~Cho, and A.~Oh, editors, {\em Advances in Neural Information Processing Systems}, volume~35, pages 25278--25294. Curran Associates, Inc., 2022.

\bibitem{mscoco}
Tsung-Yi Lin, Michael Maire, Serge Belongie, James Hays, Pietro Perona, Deva Ramanan, Piotr Doll{\'a}r, and C.~Lawrence Zitnick.
\newblock Microsoft coco: Common objects in context.
\newblock In David Fleet, Tomas Pajdla, Bernt Schiele, and Tinne Tuytelaars, editors, {\em Computer Vision -- ECCV 2014}, pages 740--755, Cham, 2014. Springer International Publishing.

\bibitem{shankar2017classificationrepresentationassessinggeodiversity}
Shreya Shankar, Yoni Halpern, Eric Breck, James Atwood, Jimbo Wilson, and D.~Sculley.
\newblock No classification without representation: Assessing geodiversity issues in open data sets for the developing world, 2017.

\bibitem{devries2019doesobjectrecognitionwork}
Terrance DeVries, Ishan Misra, Changhan Wang, and Laurens van~der Maaten.
\newblock Does object recognition work for everyone?, 2019.

\bibitem{gustafson2023pinpointingobjectrecognitionperformance}
Laura Gustafson, Megan Richards, Melissa Hall, Caner Hazirbas, Diane Bouchacourt, and Mark Ibrahim.
\newblock Pinpointing why object recognition performance degrades across income levels and geographies, 2023.

\bibitem{Radford2018ImprovingLU}
Alec Radford and Karthik Narasimhan.
\newblock Improving language understanding by generative pre-training, 2018.

\bibitem{karras2020gans}
Tero Karras, Miika Aittala, Janne Hellsten, Samuli Laine, Jaakko Lehtinen, and Timo Aila.
\newblock Training generative adversarial networks with limited data, 2020.

\bibitem{nichol2022glide}
Alex Nichol, Prafulla Dhariwal, Aditya Ramesh, Pranav Shyam, Pamela Mishkin, Bob McGrew, Ilya Sutskever, and Mark Chen.
\newblock Glide: Towards photorealistic image generation and editing with text-guided diffusion models, 2022.

\bibitem{dai2023emuenhancingimagegeneration}
Xiaoliang Dai, Ji~Hou, Chih-Yao Ma, Sam Tsai, Jialiang Wang, Rui Wang, Peizhao Zhang, Simon Vandenhende, Xiaofang Wang, Abhimanyu Dubey, Matthew Yu, Abhishek Kadian, Filip Radenovic, Dhruv Mahajan, Kunpeng Li, Yue Zhao, Vladan Petrovic, Mitesh~Kumar Singh, Simran Motwani, Yi~Wen, Yiwen Song, Roshan Sumbaly, Vignesh Ramanathan, Zijian He, Peter Vajda, and Devi Parikh.
\newblock Emu: Enhancing image generation models using photogenic needles in a haystack, 2023.

\bibitem{zhu2023minigpt4enhancingvisionlanguageunderstanding}
Deyao Zhu, Jun Chen, Xiaoqian Shen, Xiang Li, and Mohamed Elhoseiny.
\newblock Minigpt-4: Enhancing vision-language understanding with advanced large language models, 2023.

\bibitem{10017290}
Peeyush Singhal, Rahee Walambe, Sheela Ramanna, and Ketan Kotecha.
\newblock Domain adaptation: Challenges, methods, datasets, and applications.
\newblock {\em IEEE Access}, 11:6973--7020, 2023.

\bibitem{ji2024aialignmentcomprehensivesurvey}
Jiaming Ji, Tianyi Qiu, Boyuan Chen, Borong Zhang, Hantao Lou, Kaile Wang, Yawen Duan, Zhonghao He, Jiayi Zhou, Zhaowei Zhang, Fanzhi Zeng, Kwan~Yee Ng, Juntao Dai, Xuehai Pan, Aidan O'Gara, Yingshan Lei, Hua Xu, Brian Tse, Jie Fu, Stephen McAleer, Yaodong Yang, Yizhou Wang, Song-Chun Zhu, Yike Guo, and Wen Gao.
\newblock Ai alignment: A comprehensive survey, 2024.

\bibitem{wibowo2023copal}
Haryo~Akbarianto Wibowo, Erland~Hilman Fuadi, Made~Nindyatama Nityasya, Radityo~Eko Prasojo, and Alham~Fikri Aji.
\newblock Copal-id: Indonesian language reasoning with local culture and nuances.
\newblock {\em arXiv preprint arXiv:2311.01012}, 2023.

\bibitem{10.1162/tacl_a_00634}
Yong Cao, Yova Kementchedjhieva, Ruixiang Cui, Antonia Karamolegkou, Li~Zhou, Megan Dare, Lucia Donatelli, and Daniel Hershcovich.
\newblock {Cultural Adaptation of Recipes}.
\newblock {\em Transactions of the Association for Computational Linguistics}, 12:80--99, 01 2024.

\bibitem{peskov-etal-2021-adapting-entities}
Denis Peskov, Viktor Hangya, Jordan Boyd-Graber, and Alexander Fraser.
\newblock Adapting entities across languages and cultures.
\newblock In Marie-Francine Moens, Xuanjing Huang, Lucia Specia, and Scott Wen-tau Yih, editors, {\em Findings of the Association for Computational Linguistics: EMNLP 2021}, pages 3725--3750, Punta Cana, Dominican Republic, November 2021. Association for Computational Linguistics.

\bibitem{zhou-etal-2023-cross}
Li~Zhou, Laura Cabello, Yong Cao, and Daniel Hershcovich.
\newblock Cross-cultural transfer learning for {C}hinese offensive language detection.
\newblock In Sunipa Dev, Vinodkumar Prabhakaran, David Adelani, Dirk Hovy, and Luciana Benotti, editors, {\em Proceedings of the First Workshop on Cross-Cultural Considerations in NLP (C3NLP)}, pages 8--15, Dubrovnik, Croatia, May 2023. Association for Computational Linguistics.

\bibitem{zhou-etal-2023-cultural}
Li~Zhou, Antonia Karamolegkou, Wenyu Chen, and Daniel Hershcovich.
\newblock Cultural compass: Predicting transfer learning success in offensive language detection with cultural features.
\newblock In Houda Bouamor, Juan Pino, and Kalika Bali, editors, {\em Findings of the Association for Computational Linguistics: EMNLP 2023}, pages 12684--12702, Singapore, December 2023. Association for Computational Linguistics.

\bibitem{10100717}
Md~Rabiul Awal, Roy Ka-Wei Lee, Eshaan Tanwar, Tanmay Garg, and Tanmoy Chakraborty.
\newblock Model-agnostic meta-learning for multilingual hate speech detection.
\newblock {\em IEEE Transactions on Computational Social Systems}, 11(1):1086--1095, 2024.

\bibitem{hershcovich-etal-2022-challenges}
Daniel Hershcovich, Stella Frank, Heather Lent, Miryam de~Lhoneux, Mostafa Abdou, Stephanie Brandl, Emanuele Bugliarello, Laura Cabello~Piqueras, Ilias Chalkidis, Ruixiang Cui, Constanza Fierro, Katerina Margatina, Phillip Rust, and Anders S{\o}gaard.
\newblock Challenges and strategies in cross-cultural {NLP}.
\newblock In Smaranda Muresan, Preslav Nakov, and Aline Villavicencio, editors, {\em Proceedings of the 60th Annual Meeting of the Association for Computational Linguistics (Volume 1: Long Papers)}, pages 6997--7013, Dublin, Ireland, May 2022. Association for Computational Linguistics.

\bibitem{cao-etal-2024-bridging}
Yong Cao, Min Chen, and Daniel Hershcovich.
\newblock Bridging cultural nuances in dialogue agents through cultural value surveys.
\newblock In Yvette Graham and Matthew Purver, editors, {\em Findings of the Association for Computational Linguistics: EACL 2024}, pages 929--945, St. Julian{'}s, Malta, March 2024. Association for Computational Linguistics.

\bibitem{10.1145/3590773}
Federico Becattini, Pietro Bongini, Luana Bulla, Alberto~Del Bimbo, Ludovica Marinucci, Misael Mongiov\`{\i}, and Valentina Presutti.
\newblock Viscounth: A large-scale multilingual visual question answering dataset for cultural heritage.
\newblock {\em ACM Trans. Multimedia Comput. Commun. Appl.}, 19(6), jul 2023.

\bibitem{bhatia2024localconceptsuniversalsevaluating}
Mehar Bhatia, Sahithya Ravi, Aditya Chinchure, Eunjeong Hwang, and Vered Shwartz.
\newblock From local concepts to universals: Evaluating the multicultural understanding of vision-language models, 2024.

\bibitem{Katirai_2024}
Amelia Katirai, Noa Garcia, Kazuki Ide, Yuta Nakashima, and Atsuo Kishimoto.
\newblock Situating the social issues of image generation models in the model life cycle: a sociotechnical approach.
\newblock {\em AI and Ethics}, July 2024.

\bibitem{9656762}
Nassim Dehouche.
\newblock Implicit stereotypes in pre-trained classifiers.
\newblock {\em IEEE Access}, 9:167936--167947, 2021.

\bibitem{10.1145/3600211.3604711}
Ranjita Naik and Besmira Nushi.
\newblock Social biases through the text-to-image generation lens.
\newblock In {\em Proceedings of the 2023 AAAI/ACM Conference on AI, Ethics, and Society}, AIES '23, page 786–808, New York, NY, USA, 2023. Association for Computing Machinery.

\bibitem{NEURIPS2023_ae9500c4}
Yue Yu, Yuchen Zhuang, Jieyu Zhang, Yu~Meng, Alexander~J Ratner, Ranjay Krishna, Jiaming Shen, and Chao Zhang.
\newblock Large language model as attributed training data generator: A tale of diversity and bias.
\newblock In A.~Oh, T.~Naumann, A.~Globerson, K.~Saenko, M.~Hardt, and S.~Levine, editors, {\em Advances in Neural Information Processing Systems}, volume~36, pages 55734--55784. Curran Associates, Inc., 2023.

\bibitem{10.1145/3630106.3658968}
Abeba Birhane, Sepehr Dehdashtian, Vinay Prabhu, and Vishnu Boddeti.
\newblock The dark side of dataset scaling: Evaluating racial classification in multimodal models.
\newblock In {\em Proceedings of the 2024 ACM Conference on Fairness, Accountability, and Transparency}, FAccT '24, page 1229–1244, New York, NY, USA, 2024. Association for Computing Machinery.

\bibitem{Clemmer_2024_WACV}
Colton Clemmer, Junhua Ding, and Yunhe Feng.
\newblock Precisedebias: An automatic prompt engineering approach for generative ai to mitigate image demographic biases.
\newblock In {\em Proceedings of the IEEE/CVF Winter Conference on Applications of Computer Vision (WACV)}, pages 8596--8605, January 2024.

\bibitem{berg-etal-2022-prompt}
Hugo Berg, Siobhan Hall, Yash Bhalgat, Hannah Kirk, Aleksandar Shtedritski, and Max Bain.
\newblock A prompt array keeps the bias away: Debiasing vision-language models with adversarial learning.
\newblock In Yulan He, Heng Ji, Sujian Li, Yang Liu, and Chua-Hui Chang, editors, {\em Proceedings of the 2nd Conference of the Asia-Pacific Chapter of the Association for Computational Linguistics and the 12th International Joint Conference on Natural Language Processing (Volume 1: Long Papers)}, pages 806--822, Online only, November 2022. Association for Computational Linguistics.

\bibitem{10.1613/jair.1.15388}
Lukas Struppek, Dom Hintersdorf, Felix Friedrich, Manuel br, Patrick Schramowski, and Kristian Kersting.
\newblock Exploiting cultural biases via homoglyphs in text-to-image synthesis.
\newblock {\em J. Artif. Int. Res.}, 78, jan 2024.

\bibitem{Corner1980}
John Corner.
\newblock Codes and cultural analysis.
\newblock {\em Media, Culture \& Society}, 2(1), 1980.

\bibitem{GOLOUBKOV2013107}
Mikhail Goloubkov.
\newblock Literature and the russian cultural code at the beginning of the 21st century.
\newblock {\em Journal of Eurasian Studies}, 4(1):107--113, 2013.
\newblock 20 Years of the Collapse of the Fomer Soviet Union.

\bibitem{billington2010icon}
James Billington.
\newblock {\em The icon and axe: An interpretative history of Russian culture}.
\newblock Vintage, 2010.

\bibitem{figes2002natasha}
Orlando Figes.
\newblock {\em Natasha's dance: A cultural history of Russia}.
\newblock Macmillan, 2002.

\bibitem{stites1992russian}
Richard Stites.
\newblock {\em Russian popular culture: Entertainment and society since 1900}.
\newblock Cambridge University Press, 1992.

\bibitem{https://doi.org/10.1525/eth.2002.30.4.401}
Anna Wierzbicka.
\newblock Russian cultural scripts: The theory of cultural scripts and its applications.
\newblock {\em Ethos}, 30(4):401--432, 2002.

\bibitem{liu2023improved}
Haotian Liu, Chunyuan Li, Yuheng Li, and Yong~Jae Lee.
\newblock Improved baselines with visual instruction tuning, 2023.

\bibitem{NEURIPS2020_6b493230}
Patrick Lewis, Ethan Perez, Aleksandra Piktus, Fabio Petroni, Vladimir Karpukhin, Naman Goyal, Heinrich K\"{u}ttler, Mike Lewis, Wen-tau Yih, Tim Rockt\"{a}schel, Sebastian Riedel, and Douwe Kiela.
\newblock Retrieval-augmented generation for knowledge-intensive nlp tasks.
\newblock In H.~Larochelle, M.~Ranzato, R.~Hadsell, M.F. Balcan, and H.~Lin, editors, {\em Advances in Neural Information Processing Systems}, volume~33, pages 9459--9474. Curran Associates, Inc., 2020.

\bibitem{zhao2024retrievalaugmentedgenerationaigeneratedcontent}
Penghao Zhao, Hailin Zhang, Qinhan Yu, Zhengren Wang, Yunteng Geng, Fangcheng Fu, Ling Yang, Wentao Zhang, Jie Jiang, and Bin Cui.
\newblock Retrieval-augmented generation for ai-generated content: A survey, 2024.

\bibitem{arkhipkin2023fusionframesefficientarchitecturalaspects}
Vladimir Arkhipkin, Zein Shaheen, Viacheslav Vasilev, Elizaveta Dakhova, Andrey Kuznetsov, and Denis Dimitrov.
\newblock Fusionframes: Efficient architectural aspects for text-to-video generation pipeline, 2023.

\bibitem{10815947}
Vladimir Arkhipkin, Zein Shaheen, Viacheslav Vasilev, Elizaveta Dakhova, Konstantin Sobolev, Andrey Kuznetsov, and Denis Dimitrov.
\newblock Improveyourvideos: Architectural improvements for text-to-video generation pipeline.
\newblock {\em IEEE Access}, 13:1986--2003, 2025.

\bibitem{vasilev-etal-2025-ruscode}
Viacheslav Vasilev, Julia Agafonova, Nikolai Gerasimenko, Alexander Kapitanov, Polina Mikhailova, Evelina Mironova, and Denis Dimitrov.
\newblock {R}us{C}ode: {R}ussian cultural code benchmark for text-to-image generation.
\newblock In Luis Chiruzzo, Alan Ritter, and Lu~Wang, editors, {\em Findings of the Association for Computational Linguistics: NAACL 2025}, pages 7641--7657, Albuquerque, New Mexico, April 2025. Association for Computational Linguistics.

\end{thebibliography}

\section*{Appendix}

\hypertarget{appendix}{}

\begin{figure}[!ht]
    \centering
    \includegraphics[bb=0 0 10185 3740, width=0.8\linewidth]{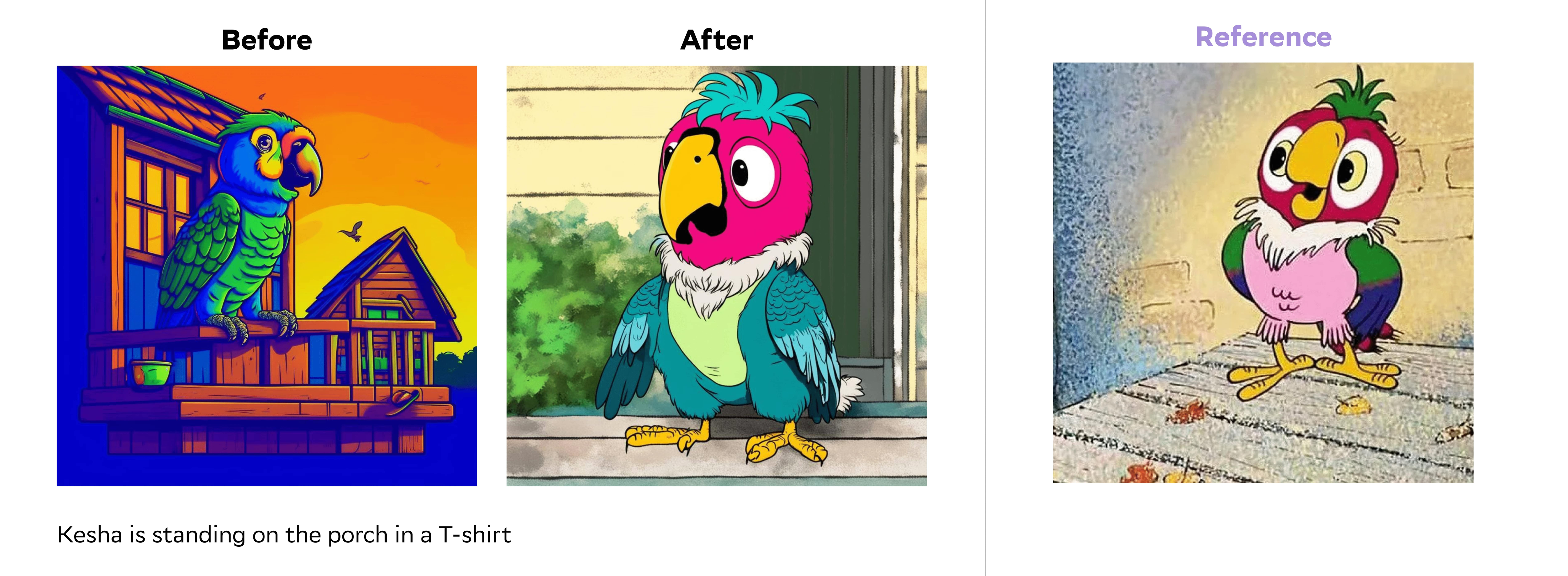}
    \caption{An example of generations by the Kandinsky 3.1 model in the Russian culture domain before and after cultural adaptation via finetuning.}
\end{figure}

\begin{figure}[!ht]
    \centering
    \includegraphics[bb=0 0 10185 3847, width=0.8\linewidth]{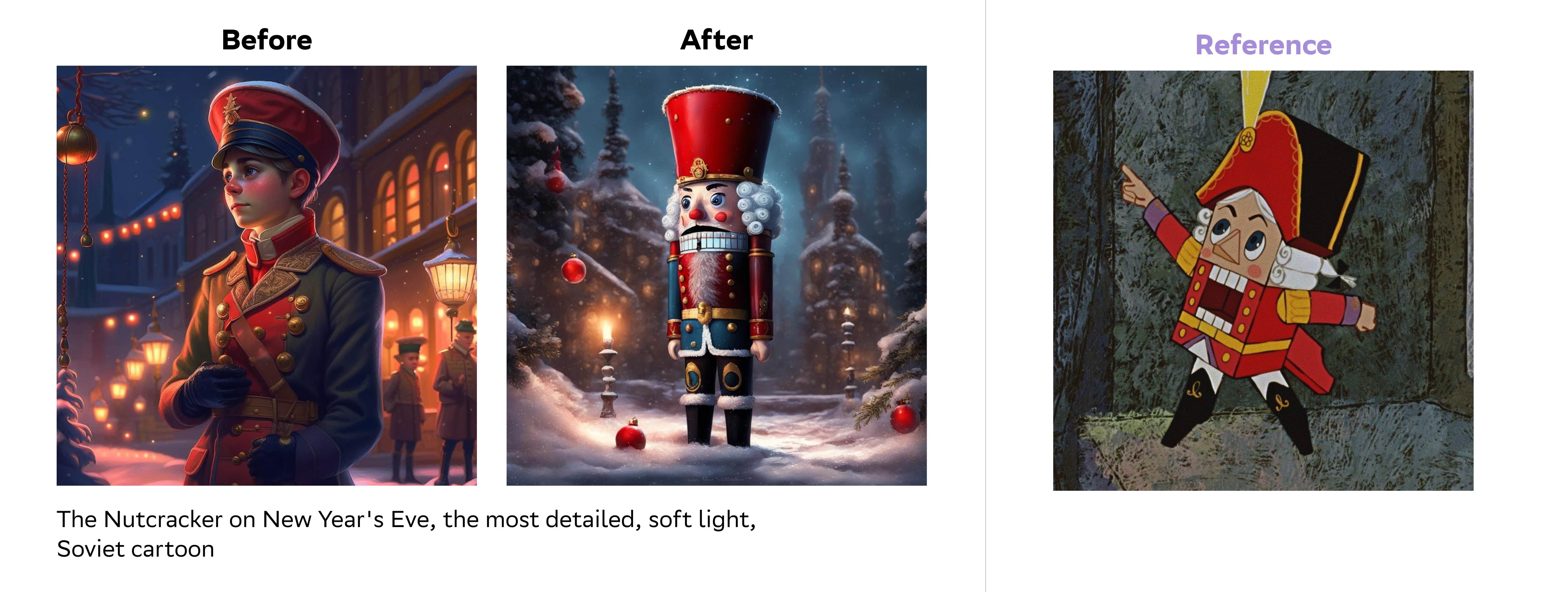}
    \caption{An example of generations by the Kandinsky 3.1 model in the Russian culture domain before and after cultural adaptation via finetuning.}
\end{figure}

\begin{figure}[!ht]
    \centering
    \includegraphics[bb=0 0 10185 3740, width=\linewidth]{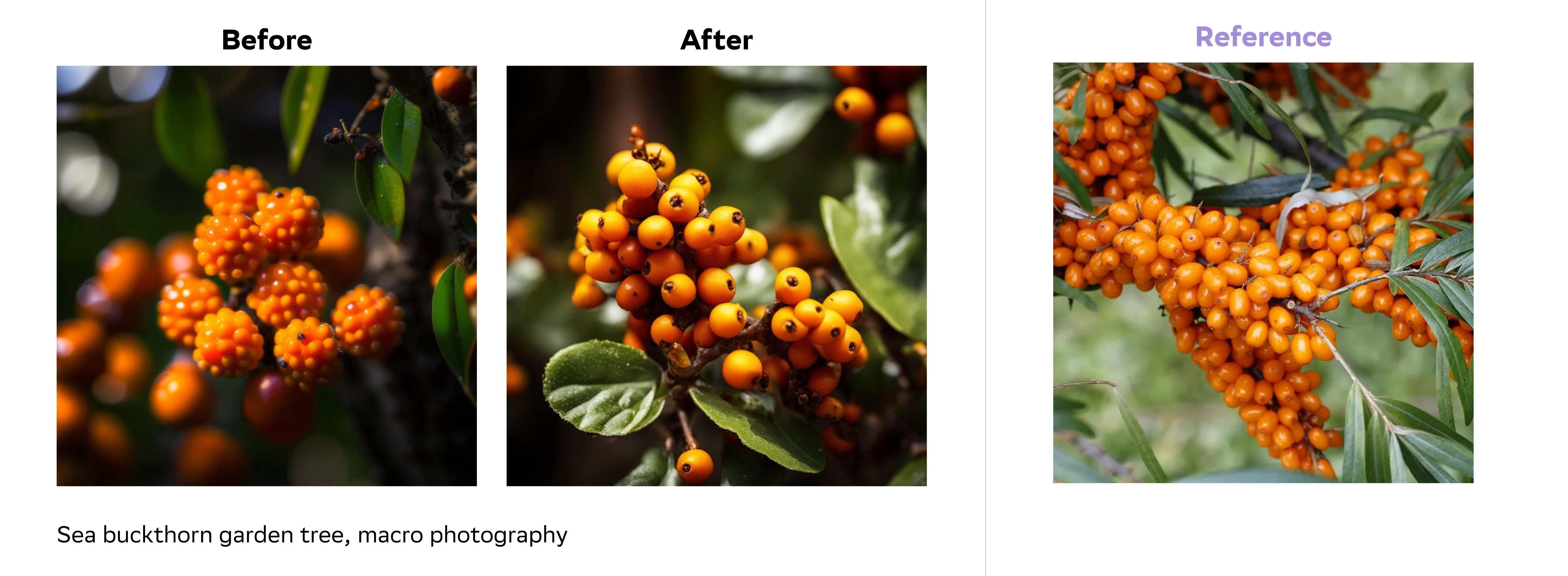}
    \caption{An example of generations by the Kandinsky 3.1 model in the Russian culture domain before and after cultural adaptation via finetuning.}
\end{figure}

\begin{figure}[!ht]
    \centering
    \includegraphics[bb=0 0 10185 3740, width=\linewidth]{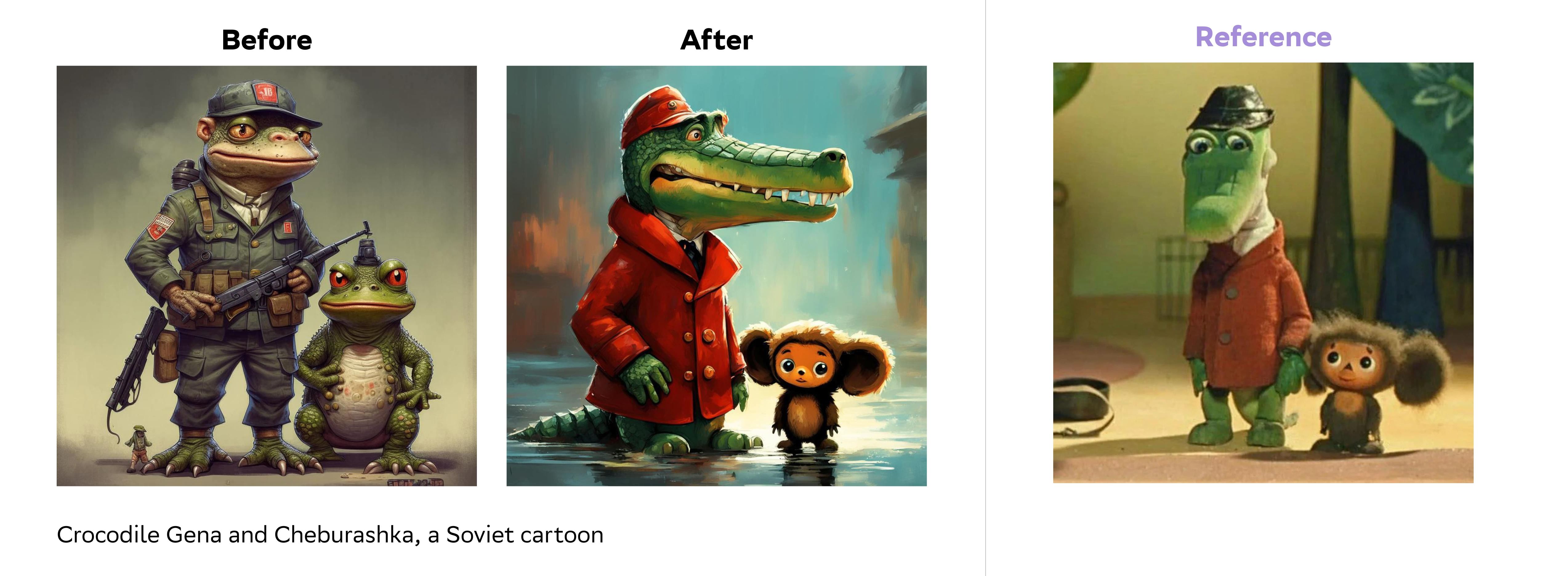}
    \caption{An example of generations by the Kandinsky 3.1 model in the Russian culture domain before and after cultural adaptation via finetuning.}
\end{figure}

\begin{figure}[!ht]
    \centering
    \includegraphics[bb=0 0 10185 3740, width=\linewidth]{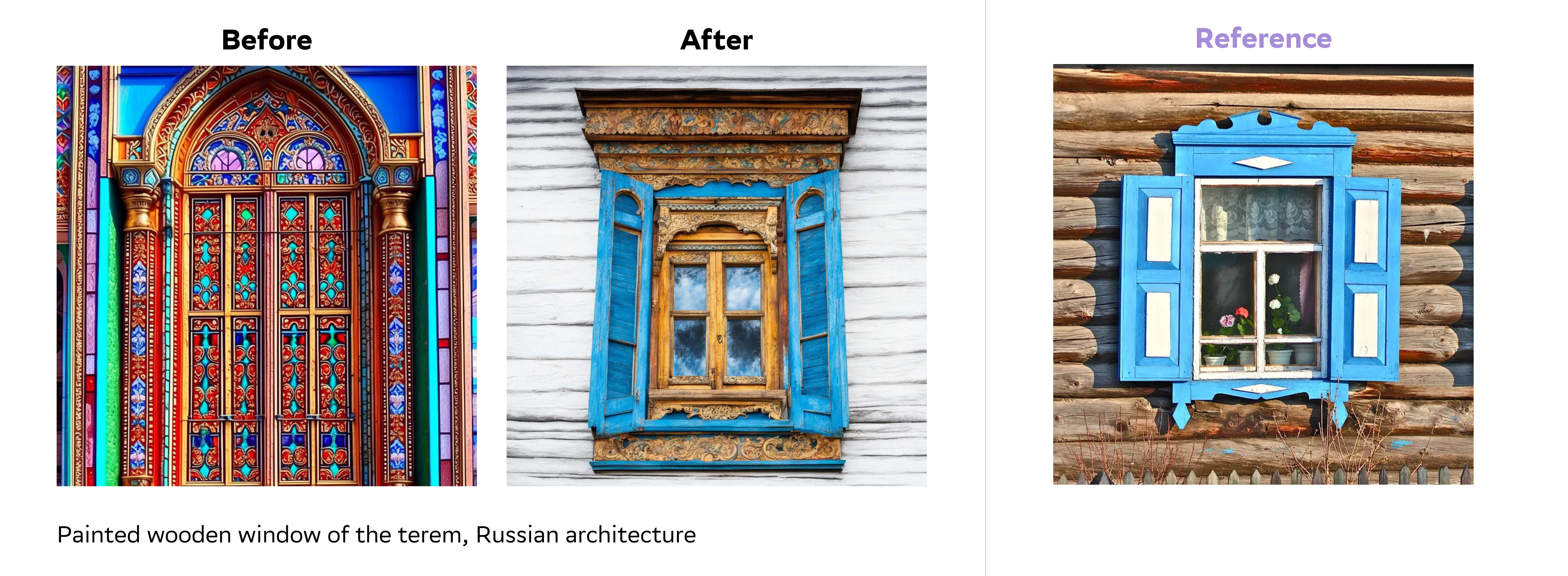}
    \caption{An example of generations by the Kandinsky 3.1 model in the Russian culture domain before and after cultural adaptation via finetuning.}
\end{figure}

\end{document}